\documentclass[11pt]{article}

\usepackage[preprint]{acl}

\usepackage{times}
\usepackage{latexsym}

\usepackage{booktabs}
\usepackage{multirow}
\usepackage{svg}
\usepackage{amsmath,amssymb}
\usepackage{here}
\usepackage{url}
\usepackage{comment}
\usepackage{multirow}
\usepackage{tabularx}
\usepackage{color}
\usepackage{xspace}
\usepackage[most]{tcolorbox}
\usepackage{xcolor}
\usepackage{pifont}

\usepackage[T1]{fontenc}
\usepackage[utf8]{inputenc}

\usepackage{microtype}

\usepackage{inconsolata}

\usepackage{graphicx}
\usepackage{colortbl}

\newcolumntype{L}[1]{>{\raggedright\let\newline\\\arraybackslash\hspace{0pt}}m{#1}}
\newcolumntype{P}[1]{>{\centering\arraybackslash}p{#1}}

\newcommand{\RowColor}{0.92}

\title{Debiasing Reward Models via Causally Motivated Inference-Time Intervention}

\author{Kazutoshi Shinoda, Kosuke Nishida, Kyosuke Nishida\\
Human Informatics Labs., NTT, Inc.\\
\texttt{kazutoshi.shinoda@ntt.com} \\
}

\begin{document}
\maketitle

\begin{abstract}
Reward models (RMs) play a central role in aligning large language models (LLMs) with human preferences. However, RMs are often sensitive to spurious features such as response length. Existing inference-time approaches for mitigating these biases typically focus exclusively on response length, resulting in performance trade-offs. In this paper, we propose causally motivated intervention for mitigating multiple types of biases in RMs at inference time. Our method first identifies neurons whose activations are strongly correlated with predefined bias attributes, and applies neuron-level intervention that suppresses these signals. We evaluate our method on RM benchmarks and observe reductions in sensitivity to spurious features across diverse bias types, without inducing performance trade-offs. Moreover, when used for preference annotation, small RMs (2B and 7B) with our method, which edits less than 2\% of all the neurons in RMs, enable LLMs to improve alignment, achieving performance comparable to that of a state-of-the-art 70B RM on AlpacaEval and MT-Bench. Further analysis reveals that bias signals are primarily encoded by neurons in early layers, shedding light on the internal mechanisms of bias exploitation in RMs.
\end{abstract}

\section{Introduction}

Reinforcement Learning from Human Feedback (RLHF) has become indispensable for aligning large language models (LLMs) with human values and preferences \cite{ouyang2022training,dubey2024llama,gemma_2024}.
Within the RLHF framework, reward models (RMs) play a pivotal role in judging the helpfulness, harmlessness, and truthfulness of LLM responses \cite{bai2022training,wang2024helpsteer,wang2025helpsteerpreference}.
RMs are typically implemented using the Bradley--Terry (BT) model \cite{bradley1952rank}, which gives a scalar score to each response.
For direct preference optimization (DPO) \cite{rafailov2023direct}, RMs are essential to determine preferred and dispreferred responses to construct preference datasets without costly manual annotation \cite{dong2024rlhf}.

\begin{figure}[t]
\small

\fbox{
\begin{minipage}{0.95\linewidth}
\textsc{Prompt} \\ [-0.4em]\rule{\linewidth}{0.4pt}
Why do leaves change color in autumn?
\end{minipage}
}

\vspace{0.5em}

\begin{minipage}{0.48\textwidth}
\fbox{
\begin{minipage}{0.95\linewidth}
\textsc{Response A (Superior format \& Incorrect)} \\[-0.5em]\rule{\linewidth}{0.4pt}
Sure! \textbf{Here is a detailed explanation about why leaves change color!}

\vspace{0.5em}
During autumn, the \textbf{temperature drops and daylight hours decrease significantly}, 
which creates many different reactions inside the leaves. Trees begin preparing for winter and 
\textbf{slow down the production of chlorophyll}, the green pigment responsible for photosynthesis.

\vspace{0.5em}
In addition, scientists have recently shown that \textbf{trees release a special ``\textit{autumn hormone}''} that triggers an internal ``color transformation process,'' causing leaves to turn bright red and purple.
\textbf{This hormone is unique to deciduous trees and does not exist during other seasons.} ...

% \vspace{0.5em}
% As chlorophyll breaks down, \textbf{other pigments become more visible}, such as carotenoids 
% and anthocyanins, contributing to the colorful scenery we enjoy every year.
% ''
\end{minipage}
}
\end{minipage}
\hfill
\begin{minipage}{0.48\textwidth}
\vspace{0.5em}
\fbox{
\begin{minipage}{0.95\linewidth}
\textsc{Response B (Standard format \& Correct)} \\[-0.5em]\rule{\linewidth}{0.4pt}
Leaves change color in autumn because plants stop producing chlorophyll as daylight decreases. 
When chlorophyll fades, other pigments such as carotenoids and anthocyanins become visible, 
creating the yellow, orange, and red colors.
\end{minipage}
}
\end{minipage}
\hfill
\begin{minipage}{0.48\textwidth}
\vspace{0.5em}
\fbox{
\begin{minipage}{0.95\linewidth}
\textsc{Reward Scores} \\[-0.6em]\rule{\linewidth}{0.4pt}
\vspace{0.1em}
\begin{tabular}{cl}
FsfairX & A(-3.0312) > B(-3.0468) \textcolor{red!70!black}{\ding{55}}\\
FsfairX + our method & A(-3.1250) < B(-3.0625) \textcolor{green!50!black}{$\checkmark$}\\
\end{tabular}
\end{minipage}
}
\end{minipage}
\vspace{-0.5em}
\caption{\textsc{Response A} has superior formatting (length, bold text, paragraphs, etc.), but it is not truthful. \textsc{Response B} is concise and truthful.
Skilled human annotators would disfavor A due to the false answer.
In contrast, the FsfairX \cite{dong2024rlhf} reward model prefer A due to its format; however, this issue is mitigated by applying our method (Figure~\ref{fig:overview}).}
\label{fig:example}
\end{figure}

\begin{figure*}[t]
    \centering
    \includegraphics[width=\textwidth]{figures/overview.png}
    \caption{Overview of our CIRM. (a) On the prepared validation set, we compute the activations of each neuron in reward models and spurious features for each bias type. Then, calculate Spearman’s $\rho$ and identify the top and bottom k neurons as bias-specific neurons (\S\ref{sec:identify}). Meanwhile, we compute the medians of the activations for each neuron. We repeat this for five types of biases introduced in \S\ref{sec:bias}. (b) At inference time, we replace the activations of the identified bias-specific neurons with their medians for five types of biases to debias rewards (\S\ref{sec:causal-intervention}).}
    \label{fig:overview}
\end{figure*}

However, RMs have been reported to suffer from biases.
The most famous one is \textit{length bias} where RMs tend to assign higher scores to longer responses regardless of their actual helpfulness \cite{singhal2024a}.
For example, \textsc{Response A} in Figure~\ref{fig:example} is factually incorrect and would be dispreferred by skilled human annotators, but it can be preferred by RMs due to the style.
Such biases pose a serious challenge in RLHF because they can propagate to LLMs through the annotation process.
When LLMs internalize the length bias, they often generate unnecessarily verbose outputs, deviating from genuine human preferences.
This issue is not limited to length but observed in other kind of stylistic cues including bold text, exclamation marks, and lists \cite{zhang-etal-2025-lists}.

To address these issues, a series of studies has proposed mitigation methods, including ensembling multiple RMs \cite{eisenstein2024helping}, averaging weights of RMs \cite{rame2024warm}, information-theoretic regularization during RM training \cite{miao2024inform}, modifying RM architectures \cite{chen2024odin} and data augmentation \cite{park-etal-2024-offsetbias}.
Despite these efforts, existing debiasing approaches typically require RM training, making them costly especially when constructing additional data or designing RM architecture for newly discovered types of biases.

Although prior work has explored inference-time debiasing for RMs \cite{dong2024rlhf,huang2024post}, existing approaches focus on coarse-grained reward adjustments based solely on response lengths, resulting in performance trade-offs between biased and unbiased evaluation settings \cite{utama-etal-2020-mind}.
Moreover, the internal mechanisms through which RMs exploit various types of biases remain underexplored.

Therefore, we propose \textbf{C}ausal \textbf{I}ntervention for \textbf{R}eward \textbf{M}odels (\textbf{CIRM}), a neuron-level inference-time method for debiasing RMs, as illustrated in Figure~\ref{fig:overview}.
In our method, we first identify neurons in RMs whose activations are highly correlated with spurious features, which we refer to as \textit{bias-specific neurons}.
We study five types of biases encompassing response length and exclamation marks to deepen the understanding of how RMs internally encode and exploit biases.
Second, we design causal intervention applied to these neurons at inference time, that constitute less than 2\% of all the neurons in RMs.
Specifically, we replace the activations of bias-specific neurons with their median values, thereby suppressing bias influence on reward predictions.

Our main contributions are three-fold:
(1) We demonstrate that our CIRM mitigates bias-induced performance degradation on unbiased subsets without inducing performance trade-offs, even though the intervened neurons constitute less than 2\% of all the neurons.
(2) We show that DPO training on preference datasets annotated using RMs with our CIRM produces LLMs with improved alignment, as evaluated on AlpacaEval 2.0, MT-Bench, and TruthfulQA. \textbf{Notably, small RMs (2--7B) equipped with our method achieve performance comparable to a large 70B RM, the current state of the art on RewardBench \cite{lambert-etal-2025-rewardbench}, on the alignment benchmarks when used for preference annotation}.
(3) We find that bias-specific neurons are predominantly located in the early layers of RMs.

\section{Method}
In this section, we first formulate standard reward models, then introduce format-related biases considered in this work, and finally present our method, CIRM.

\subsection{Preliminary}
Consider reward model $r_\theta(x)$, parameterized by $\theta$, which takes query-response pair $x$ as input and outputs a scalar reward $r$ that reflects the helpfulness of response to query.

Reward models are usually formulated with the BT model \cite{bradley1952rank}.
In the BT model, the probability that response $y_1$ is preferred to $y_2$ given query $q$ is
\begin{align}
\label{eq:bt}
p(y_1 & \succ y_2 \mid q) = \sigma(r_\theta(x_1) - r_\theta(x_2))
\end{align}
\noindent where $\sigma$ is sigmoid function.

\subsection{Bias}
\label{sec:bias}
In this work, we study five types of style-related biases as follows.
We employ these biases because RMs and natural language understanding (NLU) models are known to exploit them in previous work or are shown to use them in our experiments.
For each bias $b$, we measure the quantity of the corresponding bias $f_b(x)$ for input $x$.

\noindent\textbf{Length bias \cite{singhal2024a}}~
RMs often give high scores to long responses regardless of the actual helpfulness.
Following \citet{dong2024rlhf}, we use the character-level length of responses to quantify length bias. ($b$=len)

\noindent\textbf{Paragraph bias}~
LLMs tend to generate responses in multiple paragraphs.
To quantify this bias, we simply count the number of occurrence of ``\textbackslash n\textbackslash n'' in responses. ($b$=para)

\noindent\textbf{Overlap bias \cite{mccoy-etal-2019-right}}~
In NLU tasks, lexical overlap between two sequences is often used as shortcut cues to make predictions.
Following \citet{shinoda-etal-2023-which}, we measure overlap bias by the ratio of the common tokens contained in both query and response to the number of tokens in a response. ($b$=over)

\noindent\textbf{Exclamation mark bias \cite{zhang-etal-2025-lists}} ~
RMs can prefer exclamation marks as in ``Sure!''.
Similar to the paragraph bias, we counts the number of exclamation marks (``!'') in responses to measure exclamation mark bias. ($b$=excl)

\noindent\textbf{Bold text bias \cite{zhang-etal-2025-lists}}~
In LLM responses, bold text in Markdown is frequently used to emphasize important points, which can be preferred by RMs.
To quantify this bias, we count the number of ``**'' within responses. ($b$=bold)

\begin{figure}[tbp]
    \centering
    \begin{tabular}{cc}
        \includegraphics[width=2.8cm]{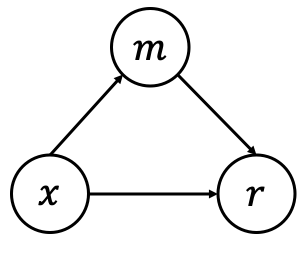} & \includegraphics[width=2.8cm]{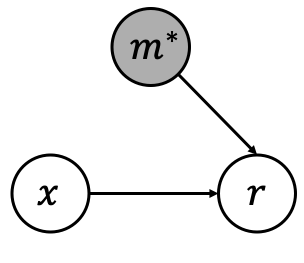} \\
        (a) Standard reward model &
        (b) Our intervention\\
    \end{tabular}
    \caption{Causal graphs for reward models.}
    \label{fig:scm}
\end{figure}

\subsection{Identifying Bias-Specific Neurons}
\label{sec:identify}
To identify neurons relatively specialized in representing the predefined biases, we compute Spearman's $\rho$ \cite{spearman} between the activations of each neuron in RMs and the quantity of each spurious feature defined in \S\ref{sec:bias}.
For simplicity, we only consider activations after the last token is input into RMs.
Following \citet{kojima-etal-2024-multilingual}, we define neurons with top and bottom $k$ Spearman's $\rho$ as bias-specific neurons.
We let $m(x)$ denote the activations of bias-specific neurons for input $x$.

\subsection{Causal Intervention}
\label{sec:causal-intervention}
To reduce the impact of the biases on rewards, we aim to answer the following counterfactual question: ``What would the rewards be if the responses were equivalent with respect to these biases?''
To answer this question, inspired by \citet{vig2020investigating}, we regard $x$, $r$, and $m$ as treatment, outcome, and mediator, respectively.
Then, we can draw a causal graph for the BT model as in Figure~\ref{fig:scm} (a), where $x$ affects $r$ directly and indirectly via $m$.

In this causal view, the BT model (Eq.~\ref{eq:bt}) can be viewed as estimating the total effect (TE) \cite{pearl2009causality}:
\begin{equation}
\label{eq:te}
\hat{{\rm TE}} = r_\theta(x_1, m(x_1)) - r_\theta(x_2, m(x_2))
\end{equation}
\noindent where we regard $m$ as input to $r_\theta$ as well.
Here, for the BT model, if TE is larger than 0, $x_1$ is judged as preferred by $r_\theta$, and vice virsa.
That is, the BT model cannot isolate the causal contribution of the content itself from the spurious features mediated by the bias-specific neurons. 

In contrast to TE, controlled direct effect (CDE) \cite{pearl2009causality} measures the effect of treatment $x$ on outcome $r$ when the mediator $m$ is fixed (controlled) to specific value $m^*$, as visualized in Figure~\ref{fig:scm} (b).
Drawing inspiration from this, we propose to predict preferences by estimating CDE:
\begin{equation}
\label{eq:cde}
\hat{{\rm CDE}} = r_\theta(x_1, m^*) - r_\theta(x_2, m^*)
\end{equation}
\noindent where we replace $m(x)$ with certain static values $m^*$.
Specifically, we set $m^*$ as the median values measured over a validation dataset $\mathcal{D}_{val}$.
Intuitively, we aim to cancel out the effect of differences in spurious features on rewards for $x_1$ and $x_2$.
We also tested replacing both $m(x_1)$ and $m(x_2)$ with zero, and replacing $m(x_1)$ with $m(x_2)$ (i.e., natural direct effect), but using medians was the best when evaluated on benchmarks for RMs.

\section{Experiments}
\label{sec:exp}

\subsection{Experimental Setups}

\noindent\textbf{Validation Set} ~
We used 500 instances randomly sampled from RewardBench as the validation set $\mathcal{D}_{val}$.
We get pairs of activations and spurious features from $\mathcal{D}_{val}$.
We also used this subset to select the hyperparamter $k$ from \{50, 100, 200, 500, 1000, 2000, 5000\} for each bias type.
Here, we jointly optimized the number of intervened neurons for each bias type, searching over the combined hyperparameter space of k across the five bias types (i.e., $7^5=16,807$ configurations).
To reduce the computational cost of hyperparameter search, we used Tree-structured Parzen Estimator \cite{bergstra2011algorithms} implemented in Optuna \cite{akiba2019optuna}.
This enables automatic adaptation of the intervention strength to each bias type while supporting the simultaneous mitigation of multiple biases.
The number of hyperparameter search trials was set to 100.
Models and datasets used in our experiments are summarized in Appendix~\ref{app:resources}.

\noindent\textbf{Baselines}~
We compared our CIRM with the following methods that do not require RM training. (1) Vanilla reward models. (2) Length penalty (LP): a naive approach to mitigate length bias. LP penalizes long reponses by adding a penalty term: $r=r_\theta(x)-\alpha|x|$. We adopted the character-level length of responses as $|x|$ and set $\alpha=0.001$ following \citet{dong2024rlhf}. (3) Locally Weighted Regression (LWR) \cite{huang2024post}: LWR assumes vanilla rewards as the sum of debiased rewards and bias terms that depend soley on lengths.
It estimates bias terms with locally weighted regression and subtracts them from vanilla rewards.
We followed the original implementation\footnote{\url{https://github.com/ZeroYuHuang/Reward-Calibration}} w.r.t. hyper parameters.

\noindent\textbf{Reward Model}~
We applied our method for two small-sized models that achieve high performance on RwardBench: FsfairX-LLaMA3-RM-v0.1 (FsfairX) \cite{dong2024rlhf}, and GRM-gemma2-2B-rewardmodel-ft (GRM) \cite{yang2024regularizing}.
These RMs employ a decoder-only transformer with a reward modeling head.
We also evaluated INF-ORM-Llama3.1-70B (INF) as a baseline, which is the state-of-the-art on RewardBench (Dec, 2025).
The RMs are summarized in Table~\ref{tab:reward-model-list}.

\noindent\textbf{Language Model}~
For alignment evaluation, we used Llama-3-8B-Instruct \cite{dubey2024llama} and Gemma-2-9b-it \cite{gemma_2024}.
All the resources used in our experiments are summarized in Appendix \ref{app:resources}.

\begin{table}[tbp]
    \small
    \centering
    \begin{tabular}{c|ccc}
    \toprule
    RM & \# Params & \# Layers & \# Neurons \\
    \midrule
    GRM & 2B & 26 & 1,246,976 \\
    FsfairX & 7B & 32 & 2,232,320 \\
    INF & 70B & 80 & 10,993,664 \\
    \bottomrule
    \end{tabular}
    \caption{Reward models used in our experiments.}
    \label{tab:reward-model-list}
\end{table}

\begin{table*}
\centering
\small
    \begin{tabular}{l|cc|cc|cc|cc|cc|c}
    \toprule
    \multicolumn{1}{c|}{RM} & $B_{\rm len}$ & $\overline{B_{\rm len}}$ & $B_{\rm para}$ & $\overline{B_{\rm para}}$ & $B_{\rm over}$ & $\overline{B_{\rm over}}$ & $B_{\rm excl}$ & $\overline{B_{\rm excl}}$ & $B_{\rm bold}$ & $\overline{B_{\rm bold}}$ & ALL \\\midrule
    GRM            & 89.94 & 89.34 & 92.79 & 90.68 & 92.20 & 89.88 & 94.33 & 85.57 & 94.00 & 100.00 & 88.96\\
    + LP & \textcolor{red!70!black}{88.14} & \textbf{90.99} & \textcolor{red!70!black}{89.47} & \textbf{92.59} & \textbf{92.77} & \textcolor{red!70!black}{89.49} & \textcolor{red!70!black}{93.62} & \textbf{87.63} & \textcolor{red!70!black}{92.00} & 100.00 & \textbf{89.13}\\
    + LWR & \textbf{90.62} & \textcolor{red!70!black}{88.35} & \textbf{93.68} & \textcolor{red!70!black}{87.50} & \textcolor{red!70!black}{91.04} & \textbf{90.66} & \textcolor{red!70!black}{93.62} & 85.57 & 94.00 & 100.00 & \textcolor{red!70!black}{88.71}\\
    + CIRM & \textbf{90.17} & \textbf{89.42} & \textbf{93.69} & \textcolor{red!70!black}{89.61} & \textcolor{red!70!black}{91.62} & \textbf{90.27} & 94.33 & 85.57 & 94.00 & 100.00 & \textbf{89.13}\\
    \midrule
    FsfairX            & 95.14 & 77.93 & 93.24 & 75.63 & 79.48 & 82.49 & 90.78 & 71.13 & 94.00 & 66.67 & 86.68\\
    + LP & \textcolor{red!70!black}{93.45} & \textbf{85.12} & \textcolor{red!70!black}{91.58} & \textbf{86.57} & \textbf{86.71} & \textbf{85.99} & \textbf{91.49} & \textbf{77.32} & \textcolor{red!70!black}{92.00} & 66.67 & \textbf{89.67}\\
    + LWR & \textcolor{red!70!black}{93.45} & \textbf{85.95} & \textcolor{red!70!black}{91.58} & \textbf{87.04} & \textbf{87.57} & \textbf{86.38} & \textbf{91.49} & \textbf{78.35} & \textcolor{red!70!black}{92.00} & 66.67 & \textbf{90.08}\\
    + CIRM & \textbf{95.25} & \textbf{78.02} & \textbf{93.69} & \textcolor{red!70!black}{74.91} & 79.48 & \textbf{83.27} & \textbf{91.49} & \textbf{72.16} & 94.00 & 66.67 & \textbf{86.80}\\
    \midrule
    INF & 96.72 & 95.70 & 97.75 & 93.91 & 97.40 & 95.72 & 97.87 & 90.72 & 98.00 & 66.67 & 96.60\\
    \bottomrule
    \end{tabular}
    \caption{Performance on RewardBench. Compared to the vanilla RMs, scores improved by LP/LWR/CIRM are shown in \textbf{bold}, while degraded scores are shown in \textcolor{red!70!black}{red}.}
\label{tab:result-reward-bench}
\end{table*}

\begin{table}
\centering
\small
    \begin{tabular}{l|cccc}
    \toprule
    \multicolumn{1}{c|}{RM} & Easy & Normal & Hard & Overall \\\midrule
    GRM    & 86.68 & 71.47 & 45.68 & 67.95\\
    + LP & \textcolor{red!70!black}{84.00} & \textbf{71.56} & \textbf{51.25} & \textbf{68.94}\\
    + LWR & \textcolor{red!70!black}{69.52} & \textcolor{red!70!black}{70.20} & \textbf{66.96} & \textbf{68.89}\\
    + CIRM & \textcolor{red!70!black}{86.18} & \textbf{71.87} & \textbf{47.59} & \textbf{68.55}\\
    \midrule
    FsfairX & 88.43 & 75.74 & 49.82 & 71.33\\
    + LP & \textcolor{red!70!black}{82.39} & \textbf{77.11} & \textbf{59.55} & \textbf{73.02}\\
    + LWR & \textcolor{red!70!black}{59.64} & \textcolor{red!70!black}{74.13} & \textbf{80.07} & \textcolor{red!70!black}{71.28}\\
    + CIRM & \textbf{88.95} & 75.74 & \textcolor{red!70!black}{49.04} & \textcolor{red!70!black}{71.24}\\
    \midrule
    INF  & 92.76 & 79.69 & 52.40 & 74.95\\
    \bottomrule
    \end{tabular}
    \caption{Performance on RM-Bench. Compared to the vanilla RMs, improved scores are shown in \textbf{bold}, while degraded scores are shown in \textcolor{red!70!black}{red}.}
    \label{tab:result-rm-bench}
\end{table}

\subsection{Performance on Reward Model Benchmarks}
\label{sec:rm-perfermance}

\noindent\textbf{Benchmark}~
We evaluated RMs on our test split of RewardBench \cite{lambert-etal-2025-rewardbench}, excluding the validation set used in \S\ref{sec:identify}, and RM-Bench \cite{liu2025rmbench}.
On these benchmarks, triplets of prompts and two responses were given, and RMs judged which response was preferred to the other.
We used accuracy as metrics.
To see if our method debias RMs, we report scores on biased subset $B_b$ and unbiased subset $\bar{B_b}$ for each bias $b$ on RewardBench.
In $B_b$ and $\bar{B_b}$, preferred response has more and fewer spurious features for bias $b$ than dispreferred response, respectively (e.g., preferred responses are longer than dispreferred ones in $B_{\rm len}$).

\noindent\textbf{Results}~
Tables~\ref{tab:result-reward-bench} and \ref{tab:result-rm-bench} report the results on RewardBench and RM-Bench, respectively.
On RewardBench, our CIRM improved accuracy on unbiased subsets for the length, overlap, and exclamation mark bias, while maintaining overall performance.
This indicates that our method effectively suppresses reliance on style-related cues without harming general capability.

Remarkably, our method not only mitigated performance degradation on $\overline{B_{\rm len}}$ but also improved scores on $B_{\rm len}$ for both GRM and FsfairX.
Existing methods for length bias, LP and LWR, often degraded scores on $B_{\rm len}$ while improving scores on $\overline{B_{\rm len}}$.
Such trade-offs between the scores on biased and unbiased subsets have been the common issue in the context of debiasing methods \cite{utama-etal-2020-mind,shinoda-etal-2021-question,shinoda-etal-2022-look}.
Given these results, neuron-level interventions would be one of the keys to solve this trade-offs.

Contrary to our expectations, for the paragraph bias, the scores on unbiased subsets are degraded by our method.
However, ablation study in Section~\ref{sec:ablation} showed intervening on paragraph-bias-specific neurons has a positive effect on downstream alignment benchmarks.
This may be due to an interplay between the intervention for different biases.
Studying this interplay is future work.

On RM-Bench, which evaluates robustness to subtle stylistic variations, existing methods consistently degraded scores on the Easy split, where stylish responses are preferred, while improving scores on the Hard split, where stylish responses are dispreferred.
In contrast, our method relatively avoided such performance degradation on the Easy split, while maintaining or improving the overall performance.
These results also imply that our CIRM does not sacrifice performance on biased data to debias reward scores.

For FsfairX, our method slightly lowered scores on the Hard split, unlike GRM.
This may be because the selected hyperparameters $k$ are different for the RMs (21k for GRM vs. 1.9k for FsfairX) as reported in Appendix~\ref{app:topk}.
If this is the case, refining the hyperparameter tuning process may help mitigate this trend.

\subsection{Performance on Alignment Benchmarks}
\label{sec:exp-alignment}
To evaluate the effect of our method on LLM alignment, we annotated preference datasets with RMs, trained LLMs with DPO \cite{rafailov2023direct} on the datasets, and evaluated the alignment.
See Appendix~\ref{app:train} for training details.

\noindent\textbf{Preference Data}~
Following \citet{huang2024post}, we annotated five responses of Llama3 and Gemma2 generated by \citet{meng2024simpo} on the Ultrafeedback \cite{cui2024ultrafeedback} prompts using RMs.
Then, we determined the response with the highest and lowest reward to be the preferred and dispreferred response, respectively.
We trained Llama and Gemma on the annotated on-policy preference datasets.
The size of the prompts is 60k.

\noindent\textbf{Benchmark}~
We evaluated alignment performance on two benchmarks: (1) AlpacaEval 2.0: pair-wise comparison against gpt-4-turbo's responses judged by gpt-4-turbo.
(2) MT-Bench: responses are rated by gpt-4o in the range from 1 to 10.
See Appendix~\ref{app:eval} for the evaluation setups.

\begin{table}[t]
\centering
\small
    \begin{tabular}{l|c|c|c|c}
    \toprule
    & \multicolumn{3}{c|}{AlpacaEval 2.0} & \\
    \multicolumn{1}{c|}{RM} & LCWR & WR & length & MT-Bench \\
    \midrule
    \multicolumn{5}{c}{\textit{Llama-3-8B-Instruct}}\\
    \midrule
    - & 26.09 & 32.06 & 1968  & 7.34\\
    \midrule
     GRM & 37.53 & 47.47 & 2193  & 7.45\\
     + LP & \textbf{44.49} & \textcolor{red!70!black}{40.18} & 1571  & \textcolor{red!70!black}{7.29}\\
     + LWR & \textbf{39.77} & \textbf{47.59} & 2119  & \textbf{7.58}\\
     + CIRM & \textbf{41.89} & \textbf{50.13} & 2201  & \textbf{7.53}\\
     \midrule
     FsfairX & 37.78 & 49.74 & 2368  & 7.64\\
     + LP & \textbf{44.03} & \textcolor{red!70!black}{46.88} & 1881  & \textcolor{red!70!black}{7.60}\\
     + LWR & \textbf{43.11} & \textcolor{red!70!black}{47.07} & 1929  & \textcolor{red!70!black}{7.44}\\
     + CIRM & \textbf{39.49} & \textbf{51.19} & 2345  & \textcolor{red!70!black}{7.62}\\
     \midrule
     INF & 40.63 & 49.61 & 2201  & 7.42\\
    \midrule
    \multicolumn{5}{c}{\textit{Gemma-2-9B-it}}\\
    \midrule
     - & 48.37 & 40.19 & 1507  & 7.83\\
     \midrule
     GRM & 58.77 & 57.24 & 1779  & 8.01\\
     + LP & \textbf{61.52} & \textcolor{red!70!black}{50.04} & 1400  & \textcolor{red!70!black}{7.84}\\
     + LWR & \textbf{59.92} & \textcolor{red!70!black}{55.55} & 1659  & \textcolor{red!70!black}{7.87}\\
     + CIRM & \textbf{59.19} & \textcolor{red!70!black}{56.62} & 1739  & \textcolor{red!70!black}{7.89}\\
     \midrule
     FsfairX & 55.32 & 58.56 & 1931  & 8.07\\
     + LP & \textbf{60.96} & \textcolor{red!70!black}{54.56} & 1611  & \textbf{8.16}\\
     + LWR & \textbf{60.60} & \textcolor{red!70!black}{51.32} & 1522  & \textcolor{red!70!black}{7.96}\\
     + CIRM & \textbf{57.76} & \textbf{60.52} & 1923  & \textbf{8.15}\\
     \midrule
     INF & 58.98 & 61.51 & 1919  & 8.11\\
    \bottomrule
    \end{tabular}
    \caption{Evaluation of LLM alignment on AlpacaEval2.0 and MT-Bench. Compared to the vanilla RMs, improved scores are shown in \textbf{bold}, while degraded scores are shown in \textcolor{red!70!black}{red}. (LCWR: length-controlled win rate, WR: win rate)}
    \label{tab:result-alignment}
\end{table}

\begin{table}[tbp]
\centering
\small
    \begin{tabular}{l|c|c|c|c}
    \toprule
    & \multicolumn{3}{c|}{AlpacaEval 2.0} & \\
    \multicolumn{1}{c|}{Method} & LCWR & WR & length & MT-Bench \\
    \midrule
    CIRM & \underline{39.49} & \textbf{51.19} & 2345 & \underline{7.62}\\
    \midrule
    $-$ len & 37.10 & 49.83 & 2370  & \textbf{7.81}\\
    $-$ para & 38.20 & \underline{50.89} & 2370  & 7.53\\
    $-$ over & \textbf{40.02} & 50.24 & 2359  & 7.54\\
    $-$ excl & 37.06 & 50.21 & 2364  & 7.56\\
    $-$ bold & 38.38 & 50.06 & 2371  & 7.29\\
    \bottomrule
    \end{tabular}
    \caption{Ablation study for mitigated biases using our method with Llama3-8B and FsfairX. The best score is shown in \textbf{bold}, and the second-best score is \underline{underlined}.}
    \label{tab:result-ablation}
\end{table}

\noindent\textbf{Results}~
Table~\ref{tab:result-alignment} presents the results.
Applying our method consistently lead to improved alignment metrics without causing trade-offs between evaluation metrics, mirroring the trend observed on RM benchmarks.
Our intervention improved both the length-controlled win rate (LCWR) \cite{dubois2024lengthcontrolled}, which reduces gameability due to response length, and win rate (WR) on AlpacaEval in most cases.
Our method maintained or improved MT-Bench scores as well.
\textbf{Notably, the small RMs equipped with our CIRM enabled DPO-trained LLMs to achieve performance comparable to that of LLMs trained with annotations from the 70B RM on the benchmarks}.

On the other hand, existing debiasing methods, LP and LWR, tended to degrade WR and MT-Bench scores.
This may be because existing methods penalized longer responses more than necessary, which also lead to the performance degradation in the biased subset in RM benchmarks.
These results suggest that performance trade-offs observed in RM benchmarks can propagate to trade-offs in downstream alignment performance.

In addition, the average accuracies on RM benchmarks were not correlated with alignment scores.
E.g., INF achieved the best on the two RM benchmarks, but it often ragged behind other RMs equipped with debiasing methods on the alignment benchmarks.
This finding highlights the importance of evaluating robustness to bias when assessing the performance of RMs.

\section{Analysis}
\label{sec:analysis}

\subsection{Is it necessary to consider the five types of biases simultaneously?}
\label{sec:ablation}
Table~\ref{tab:result-ablation} shows the result of ablation study for bias mitigation.
The result shows that when mitigating the five types of biases using the proposed method, it most effectively resolves the performance trade-offs between AlpacaEval and MT-Bench after alignment, i.e., ours is second best in LCWR, best in WR, and second best in MT-Bench.

\subsection{Where Do Bias-Specific Neurons Exist in Reward Models?}
We examined where bias-specific neurons are located within the reward models. Figures~\ref{fig:spearman-layer-grm} and~\ref{fig:spearman-layer-fsfairx} show the layer-wise distribution of neurons with the top and bottom 500 Spearman’s $\rho$ for each bias type. Across GRM and FsfairX, these neurons are not evenly distributed but appear predominantly in early layers in most cases, suggesting that spurious features are captured through shallow feature transformations rather than deeper semantic processing.
One exception is the overlap bias in FsfairX, which may stem from the use of a summarization dataset \cite{dong2024rlhf} that naturally contains substantial lexical overlap between prompts and responses during RM training.

\def\WidHist{2.5cm}
\begin{figure*}[t]
\small
    \begin{tabular}{ccccc}
    Length & Paragraph & Overlap & Excl. Mark & Bold \\
    \includegraphics[width=\WidHist]{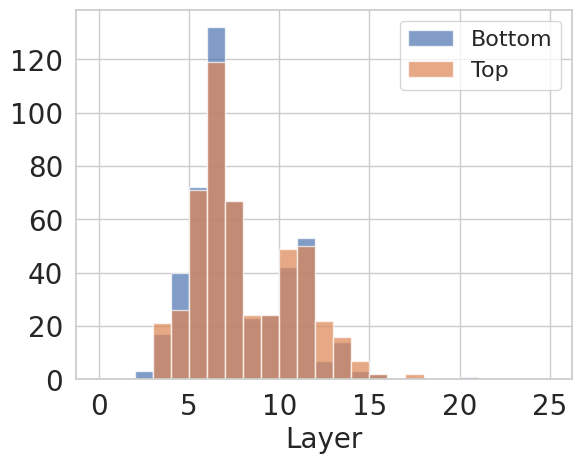} & \includegraphics[width=\WidHist]{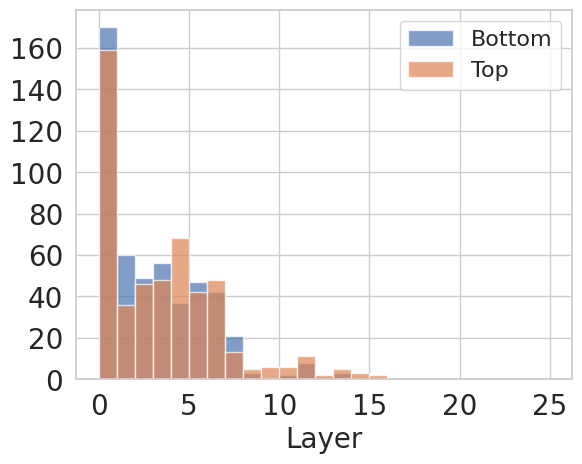} & \includegraphics[width=\WidHist]{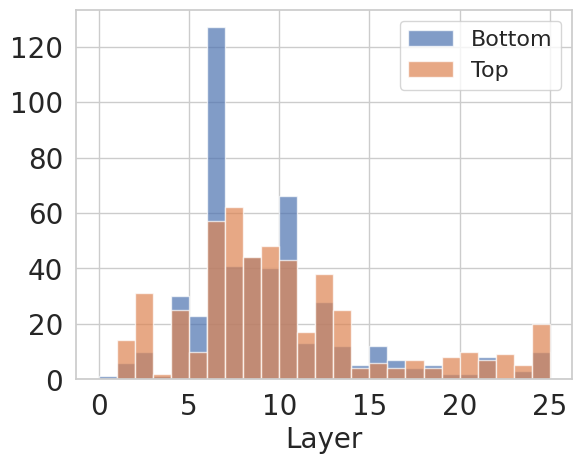} & \includegraphics[width=\WidHist]{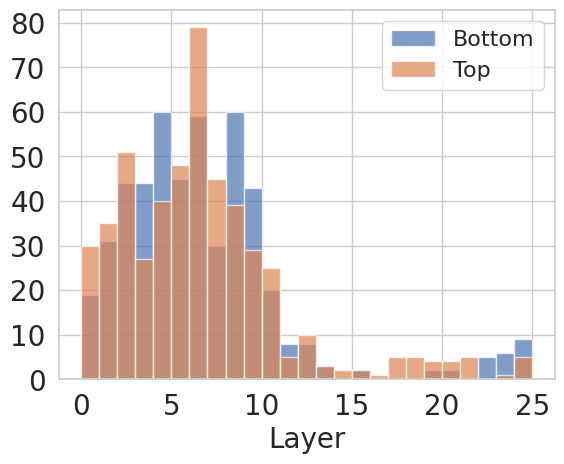} & \includegraphics[width=\WidHist]{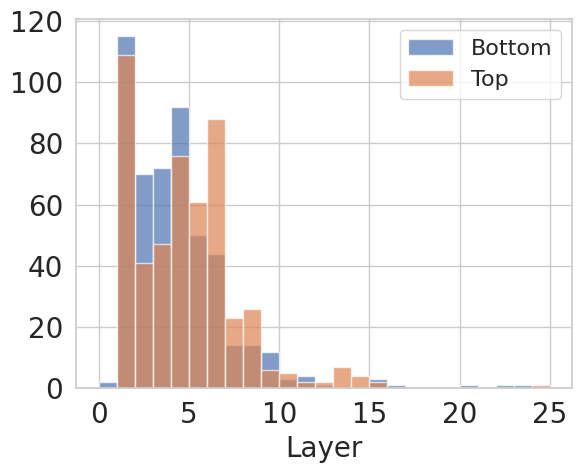} \\
    \end{tabular}
\caption{Histogram of neurons with the top and bottom 500 Spearman’s $\rho$ of GRM across layers.}
\label{fig:spearman-layer-grm}
\end{figure*}

\def\WidHist{2.5cm}
\begin{figure*}[t]
\small
    \begin{tabular}{ccccc}
    Length & Paragraph & Overlap & Excl. Mark & Bold \\
    \includegraphics[width=\WidHist]{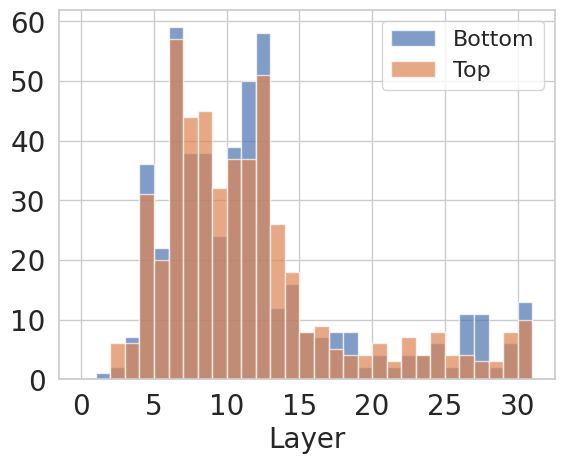} & \includegraphics[width=\WidHist]{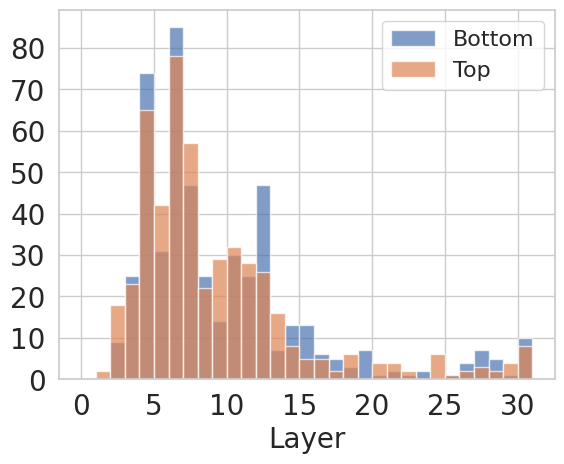} & \includegraphics[width=\WidHist]{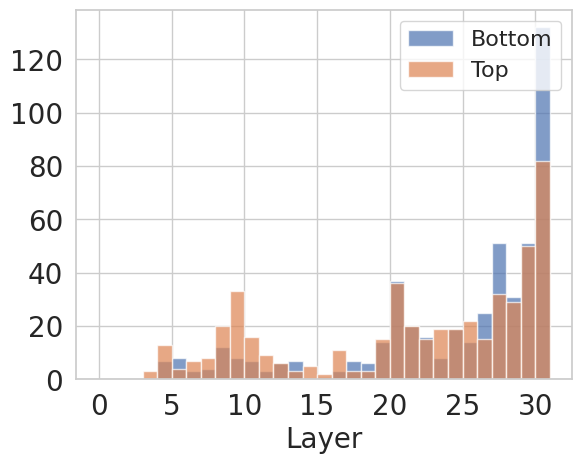} & \includegraphics[width=\WidHist]{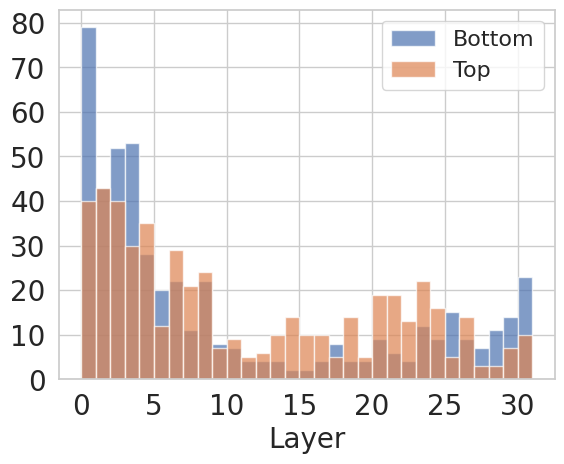} & \includegraphics[width=\WidHist]{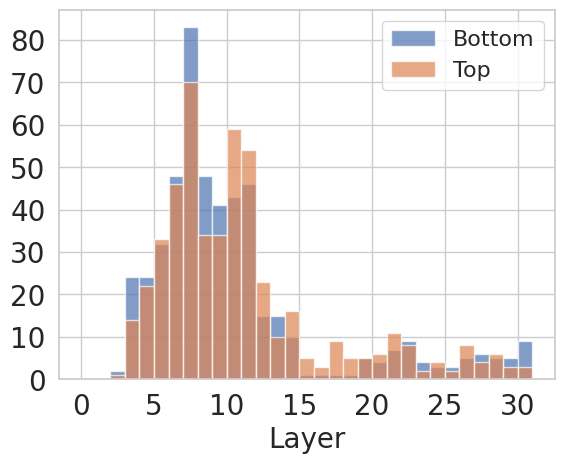} \\
    \end{tabular}
\caption{Histogram of neurons with the top and bottom 500 Spearman’s $\rho$ of FsfairX across layers.}
\label{fig:spearman-layer-fsfairx}
\end{figure*}

Figures~\ref{fig:dist-2d-neuron-grm} and~\ref{fig:dist-2d-neuron-fsfairx} in Appendix~\ref{app:dist-neurons} further decompose these neurons by architectural component, showing that most bias-specific neurons exist in query projection, up projection, and gate projection layers.
For length bias, most bias-specific neurons exist in query projection layers.
This is reasonable because only query vectors in self-attention at the last token can attend to the whole context in decoder-only transformers.
For other biases, bias-specific neurons are often found in up and gate projection layers.
That is, the two layers may capture token-level superficial cues.
This finding is consistent with \citet{meng2022locating}'s hypothesis that up projection layers retrieve knowledge stored in down projection layers.
Overall, bias-specific neurons occupy specific regions of RMs, providing a structural basis for why targeted neuron-level interventions can effectively modulate behaviors without degrading the overall performance.

\subsection{Does Our Method Debias Preference Datasets?}
We examined whether our method reduces the extent to which RMs favor biased responses during annotation in Section \ref{sec:exp-alignment}.
Table~\ref{tb:bias-ratio} reports the proportion of preferred responses that exhibit higher spurious features in datasets annotated.
For GRM, applying the intervention consistently lowers the ratio of biased preferences.
These results confirm that replacing the activations of bias-specific neurons with their median values successfully suppresses bias-driven preferences at annotation time.
For FsfairX, the ratios were not chancged significantly.
This would be because the number of intervened neurons were fewer than GRM as in Appendix \ref{app:topk}.
For LP and LWR, the ratios were reduced by 10-20 \% points for length and paragraph biases.
This excessive suppression likely caused the degradation in WR and MTbench scores in Section~\ref{sec:exp-alignment}.

\begin{table}[t]
    \centering
    \small
    \begin{tabular}{l|ccccc}
    \toprule
    \multicolumn{1}{c|}{RM} & len & para & over & excl & bold \\
     \midrule
     GRM & 54.85 & 55.88 & 55.51 & 67.61 & 65.25\\
     + LP & 25.65 & 39.06 & 51.98 & 59.02 & 55.44\\
     + LWR & 52.70 & 55.20 & 55.84 & 66.27 & 63.50\\
     + CIRM & 51.71 & 52.14 & 55.99 & 61.90 & 64.97\\
     \midrule
     FsfairX & 63.19 & 57.44 & 54.56 & 51.95 & 67.13\\
     + LP & 41.11 & 46.33 & 53.92 & 48.67 & 59.33\\
     + LWR & 41.66 & 44.11 & 53.38 & 47.59 & 61.25\\
      + CIRM & 63.35 & 57.75 & 54.24 & 53.02 & 68.27\\
    \midrule
     INF & 50.75 & 55.38 & 56.03 & 65.01 & 66.79\\
    \bottomrule
    \end{tabular}
    \caption{Ratio (\%) of the biased subset, $|B|/(|B|+|\overline{B}|
    )$, in annotated preference data for each bias.
    }
    \label{tb:bias-ratio}
\end{table}

\begin{table}[t]
    \centering
    \small
    \begin{tabular}{l|ccccc}
\toprule
\multicolumn{1}{c|}{RM} & len & para & over & excl & bold\\
\midrule
 - & 1323.09 & 7.80 & 0.27 & 0.42 & 14.82\\
 \midrule
GRM & 1538.21 & 8.62 & 0.27 & 0.50 & 21.22\\
+ LP & 1194.45 & 7.64 & 0.28 & 0.66 & 18.30\\
+ LWR & 1446.14 & 8.40 & 0.28 & 0.62 & 19.98\\
+ CIRM & 1511.32 & 7.85 & 0.28 & 0.49 & 20.12\\
\midrule
FsfairX & 1612.35 & 7.85 & 0.29 & 0.24 & 18.64\\
+ LP & 1346.14 & 6.65 & 0.29 & 0.24 & 15.25\\
+ LWR & 1289.46 & 6.20 & 0.30 & 0.20 & 14.20\\
+ CIRM & 1610.51 & 7.90 & 0.29 & 0.25 & 18.38\\
\midrule
INF & 1668.78 & 8.06 & 0.29 & 0.39 & 19.77\\
\bottomrule
    \end{tabular}
    \caption{Mean values of spurious features in LLM responses. ``-'': Gemma2 without additional DPO training.}
    \label{tab:response}
\end{table}

\subsection{Are Biases in LLM Responses Reduced After DPO with Our Method?}
We analyzed whether debiasing RMs during annotation leads to less biased generations in LLMs after DPO.
Table~\ref{tab:response} shows the mean spurious feature values in Gemma2’s responses to the MT-Bench questions.
DPO training with vanilla RMs amplifies several biases, especially response length as commonly observed in DPO \cite{park-etal-2024-disentangling}.

When using LP or LWR, the response lengths were decreased significantly compared to vanilla RMs.
However, the lengths can be shorter than even Gemma2 without additional DPO training, which was also observed in AlpacaEval (Table~\ref{tab:result-alignment}).
In addition, we found that these methods amplified excl. mark bias for GRM from 0.5 to 0.6.

In contrast, with our method, the decrease in spurious features were more moderate compared to LP and LWR.
For example, for GRM, the average response length decreases from 1538.21 to 1511.32 tokens under our CIRM.
Unlike LP and LWR, our CIRM did not cause bias amplification.
These results suggest that our neuron-level intervention edited only the necessary neurons, thereby avoiding significant adverse effects in LLM alignment.
Qualitative analyses in Appendix~\ref{app:examples} also supports this claim.

\subsection{Does Our Method Improve the Truthfulness of LLMs?}
If our method can successfully mitigate format-related biases, the annotated preference datasets will focus on response content rather than format, which may improve truthfulness in aligned LLMs.
To verify this hypothesis, we evaluate LLMs on TruthfulQA \cite{lin-etal-2022-truthfulqa}, a widely used benchmark that probes models’ tendency to generate false answers under misleading premises.
See Appendix~\ref{app:eval} for evaluation details.

Table \ref{tab:truthful} shows the results.
In three of the four settings, our CIRM improves LLM truthfulness compared to vanilla RMs.
These results suggest that our CIRM not only reduces biases in preference annotation but also better reflects factual correctness.

\begin{table}[t]
    \centering
    \small
    \begin{tabular}{l|c}
    \toprule
    \multicolumn{1}{c|}{RM} & TruthfulQA\\
    \midrule
    \multicolumn{2}{c}{\textit{Llama-3-8B-Instruct}}\\
    \midrule
    GRM & 58.50 \\
    + CIRM & \textbf{59.50} \\
    \midrule
    FsfairX & 59.21 \\
    + CIRM & \textcolor{red!70!black}{58.89} \\
    \midrule
    \multicolumn{2}{c}{\textit{Gemma-2-9B-it}}\\
    \midrule
    GRM & 62.78 \\
    + CIRM & \textbf{63.06} \\
    \midrule
    FsfairX & 59.99 \\
    + CIRM & \textbf{60.39} \\
    \bottomrule
    \end{tabular}
    \caption{Performance on TruthfulQA. Compared to the vanilla RMs, improved scores are shown in \textbf{bold}, while degraded scores are shown in \textcolor{red!70!black}{red}.}
    \label{tab:truthful}
\end{table}

\begin{comment}
\section{Related Work}
It is well-known that RMs can exploit length bias \cite{singhal2024a}, which results in LLMs that generate unnecessarily long responses.
Length bias has been addressed by 

 \citet{park-etal-2024-offsetbias} defined multiple types of biases in preference data and proposed a data augmentation approach to mitigate them.

 \citet{shinoda-etal-2023-which} showed that single token or phrases can be easily used as shortcuts by language models.
\end{comment}

\section{Related Work}
\subsection{Biases in Reward Models}
Several studies have shown that RMs exploit superficial features.
\citet{zhang-etal-2025-lists} show that format bias (lists, links, bold text, emojis) shifts RM preferences.
\citet{park-etal-2024-offsetbias} demonstrate that targeted data augmentation improves robustness against multiple types of biases.
\citet{chen2024odin} proposes adding heads to RMs to disentangle lengths from rewards.
InfoRM introduces a variational information bottleneck for reward modeling \cite{miao2024inform}.
These methods, while effective, require re-training or architectural changes.
Complementary to training-time approaches, length penalty has been employed to penalize long responses \cite{dong2024rlhf}.
\citet{huang2024post} present post-hoc reward calibration that estimates and removes a bias term that relies solely on length.
In contrast, our approach introduces causally motivated test-time intervention that directly targets RM neurons where multiple bias signals propagate, thereby improving the quality of preference annotations without inducing performance trade-offs.
In addition, our neuron-level analysis also reveals where such biases are encoded in RMs across layers and components, providing an interpretability benefit that complements the intervention itself.

\subsection{Neuron-level Intervention}
Neuron-level intervention methods have been used to identify internal mediators of model predictions, enabling more targeted interventions than gradient-based attribution \cite{pmlr-v70-sundararajan17a} or probing methods \cite{alain2016understanding}.
These methods modify specific activations to alter model behavior while leaving most of the computation unchanged.
\citet{vig2020investigating} used this framework to study gender bias in language models, theoretically grounded in causal mediation analysis \cite{pearl2009causality}.
\citet{meng2022locating} located and edited knowledge in LLMs.
\citet{kojima-etal-2024-multilingual} identified language-specific neurons to control the response language of LLMs.

While prior studies typically target a single attribute (e.g., gender bias), our work addresses multiple types of format-related biases in RMs for LLM alignment.
Unlike existing neuron editing approaches, we demonstrate effectiveness not only on RM benchmarks but also on downstream alignment performance, highlighting the practical impact of our method on RLHF.
Moreover, to the best of our knowledge, this is the first work to reinterpret the Bradley--Terry reward model as estimating the total effect (TE) and to derive a debiasing intervention based on the controlled direct effect (CDE).

\section{Conclusion}
We introduced an inference-time method for debiasing RMs using causally motivated intervention on bias-specific neurons.
By identifying neurons that strongly correlate with five types of commonly observed spurious features, and replacing their activations with median values, our approach effectively reduces the influence of spurious features without training.
Experimental results show that this intervention both enhances robustness on RM benchmarks and produces debiased preference annotations for downstream DPO training.
In addition, our method avoided causing performance trade-offs on RM and alignment benchmarks compared to existing test-time debiasing methods.
Our analysis supports that our CIRM debias RMs and preference datasets, thereby improving truthfulness of aligned LLMs.

Future work includes exploring more fine-grained causal structures within RMs, extending the intervention to other biases such as list, and applying similar causal techniques to other components of the RLHF pipeline such as preference datasets or policy models.
Overall, our results position inference-time causal intervention as a lightweight, generalizable, and model-agnostic tool for improving the reliability of RMs and the alignment of LLMs.

\section*{Limitations}
Our work focuses on five predefined bias types (length, paragraph, overlap, exclamation mark, bold text).
It's unclear how well this method generalizes to emergent biases beyond these types.
In addition, the identification of bias-specific neurons depends on a small validation set and requires hyperparameter tuning, which introduces potential for overfitting.
The method also relies on a simple causal graph, which may be less effective when bias signals are broadly distributed or not captured by the mediator.
Finally, our evaluation of LLM alignment relies on LLM-as-a-judge metrics, which may suffer from biases similar to the models being evaluated.

\bibliography{custom}

\appendix

\section{Hyperparameter Tuning}
\label{app:topk}
To determine bias-specific neurons, we tune the hyperparameter $k$ for each bias as described in Section~\ref{sec:exp}.
The result of hyperparameter tuning is in Table~\ref{tab:hyper-topk}.
Based on this, the total neurons intervened by our method constitute 1.7\% (= 10,750 * 2 / 1,246,976) and 0.085\% (= 950 * 2 / 2,232,320) for GRM and FsfiarX, respectively.

\begin{table}[h]
    \centering
    \small
    \begin{tabular}{c|c|c}
    \toprule
    $b$ & GRM & FsfairX \\
    \midrule
    len & 5000 & 500 \\
    para & 5000 & 100 \\
    over & 500 &  100 \\
    excl & 200 & 50 \\
    bold & 50 &  200 \\
    \midrule
    SUM & 10,750 & 950 \\
    \bottomrule
    \end{tabular}
    \caption{Hyperparameter (top and bottom $k$ of Spearman's $\rho$) for determining bias-specific neurons.}
    \label{tab:hyper-topk}
\end{table}

\section{Training Details}
\label{app:train}
Hyperparameters for DPO training are given in Table~\ref{tab:hyper-dpo}.
We compared 32 and 64 for batch size and found 64 was better for Llama-3-8B-Instruct, so we set the batch size as 64 for every DPO run.
We set the other parameters following \citet{meng2024simpo}.

\begin{table}[h]
    \centering
    \small
    \begin{tabular}{c|c}
    \toprule
    Hyperparameter & Value \\
    \midrule
    $\beta$ & 0.1 \\
    epoch & 1 \\
    learning rate & 5e-7 \\
    learning rate schedule & cosine \\
    warmup ratio & 0.1 \\
    batch size & 64 \\
    max length & 2048 \\
    \bottomrule
    \end{tabular}
    \caption{Hyperparameters for DPO training.}
    \label{tab:hyper-dpo}
\end{table}

\section{Evaluation Details}
\label{app:eval}

\subsection{Reward Model Evaluation}
For RewardBench, we used the ``filtered'' subset, which is provided in \href{https://huggingface.co/datasets/allenai/reward-bench/viewer/default/filtered}{this link}.

For RM-Bench, we followed the official score calculation process (Easy, Normal, Hard, and Overall) averaging over different domains as in \href{https://github.com/THU-KEG/RM-Bench}{this code}.

\subsection{Alignment Evaluation}

Hyperparameters we used for generating responses in alignment evaluation on AlpacaEval 2.0 and MTBench are in Table~\ref{tab:hyper-alignment}.

\begin{table}[h]
    \centering
    \small
    \begin{tabular}{c|c|P{2.3cm}}
    \toprule
    Hyperparameter & AlpacaEval & MTBench \\
    \midrule
    do\_sample & True & True \\
    max\_new\_tokens & 2048 & 2048 \\
    top\_p & 1.0 & 1.0 \\
    temperature & 0.7 & different values defined for different categories\\
    \bottomrule
    \end{tabular}
    \caption{Hyperparameters of response generation for alignment evaluation.}
    \label{tab:hyper-alignment}
\end{table}

For AlpacaEval 2.0, we used the evaluator settings provided in \href{https://github.com/tatsu-lab/alpaca_eval/blob/main/src/alpaca_eval/evaluators_configs/weighted_alpaca_eval_gpt4_turbo_new/configs.yaml}{this link}.
When evaluating alignment on MT-Bench, we generally followed \href{https://github.com/lm-sys/FastChat/tree/main/fastchat/llm_judge#mt-bench}{this code}, but adopted GPT-4o as a judge due to the high cost of GPT-4 used in the original setting.

For TruthfulQA, we conduct zero-shot evaluation using the Language Model Evaluation Harness \cite{eval-harness} and report performance with the MC2 metric.

\section{Distributions of Bias-Specific Neurons}
\label{app:dist-neurons}
Figures~\ref{fig:dist-2d-neuron-fsfairx} and \ref{fig:dist-2d-neuron-grm} show the distributions of neurons with top and bottom 500 Spearman's $\rho$ across layers and model components.

\section{Qualitative Analysis}
\label{app:examples}
In this section, we present examples of responses generated by Gemma2 and Llama3 before and after DPO training with annotations with and without our method.
We take the examples from Alpaca Eval.
Tables~\ref{tab:example-llama3} and \ref{tab:example-gemma2} show the examples generated by Llama3 and Gemma2, respectively.
The following trends are found in these examples.

\paragraph{DPO with vanilla GRM amplifies length, bold text, and paragraph biases.}
For both LLMs, the length, number of ``**'' and ``\textbackslash n\textbackslash n'' are increased via DPO with vanilla GRM.
These qualitative examples support that vanilla RMs can amplify these biases significantly.
Especially for Gemma2, DPO with GRM annotations seems to increase the number of paragraphs than necessarary, e.g., ``**Philanthropy:**\textbackslash n\textbackslash n'' and ``**Current Status:**\textbackslash n\textbackslash n''.

\paragraph{Our method mitigated the bias amplification in DPO.}
In both cases, our method does not amplify the biases significantly.
For instance, the number of paragraphs increase from five to six for Gemma2, and does not change for Llama3 via DPO when using our method.
In addition, the use frequency of bold text is not increased in Llama3's responses.

\paragraph{Our method can mitigate unknown biases.}
Although list structures (e.g., ``* ...") are not explicitly considered in our method, it effectively suppresses the unnecessary use of lists in Gemma2’s responses, unlike the vanilla GRM.
This observation suggests that our method can generalize to unknown biases that are not explicitly accounted for.
Exploring the generalization capability of our method to unknown biases is left for future work.

\section{Resources}
\label{app:resources}
The links to all the models and datasets we used in our experiments are summarized in Table~\ref{tab:resources}.

\clearpage

\def\WidRatio{0.7}
\begin{figure}[tb]
\centering
\small
Length\\
\includegraphics[width=\WidRatio\linewidth]{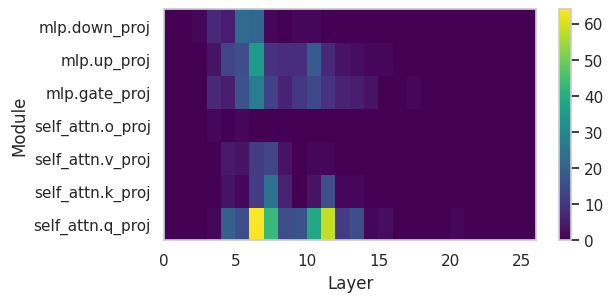}\\
Paragraph\\
\includegraphics[width=\WidRatio\linewidth]{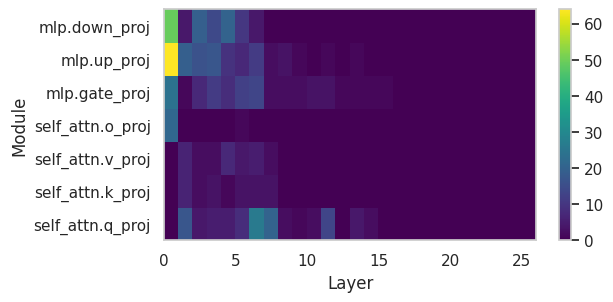}\\
Overlap\\
\includegraphics[width=\WidRatio\linewidth]{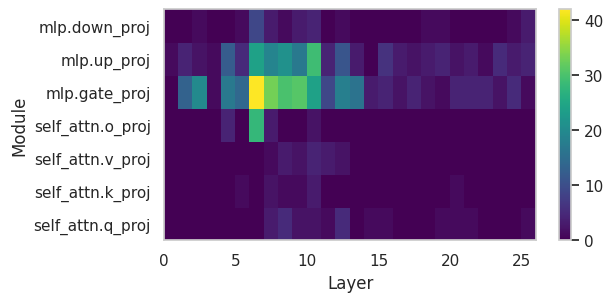}\\
Exclamation\\
\includegraphics[width=\WidRatio\linewidth]{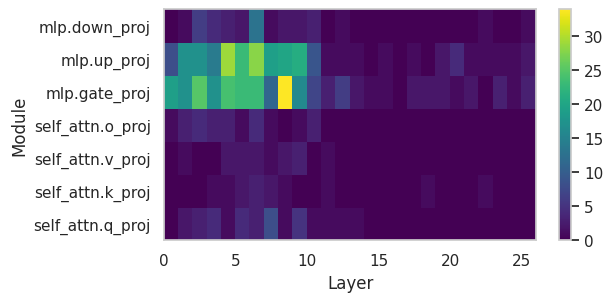}\\
Bold \\
\includegraphics[width=\WidRatio\linewidth]{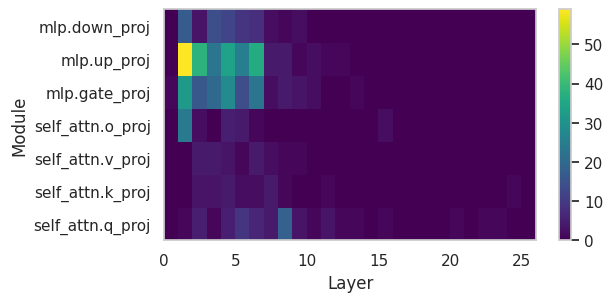}\\
\caption{Distributions of neurons in GRM with top and bottom 500 Spearman's $\rho$.}
\label{fig:dist-2d-neuron-grm}
\end{figure}

\def\WidRatio{0.7}
\begin{figure}[t]
\centering
\small
Length\\
\includegraphics[width=\WidRatio\linewidth]{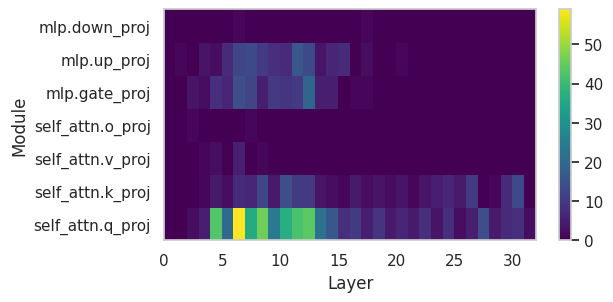}\\
Paragraph\\
\includegraphics[width=\WidRatio\linewidth]{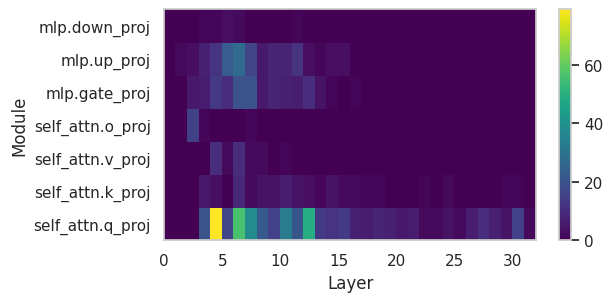}\\
Overlap\\
\includegraphics[width=\WidRatio\linewidth]{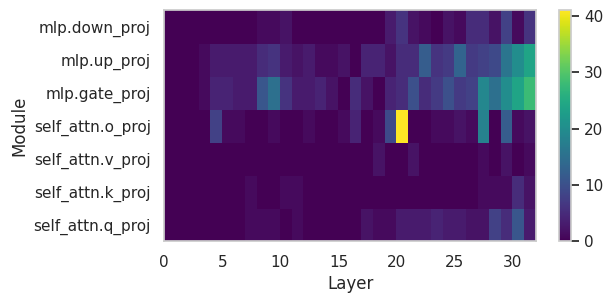}\\
Exclamation\\
\includegraphics[width=\WidRatio\linewidth]{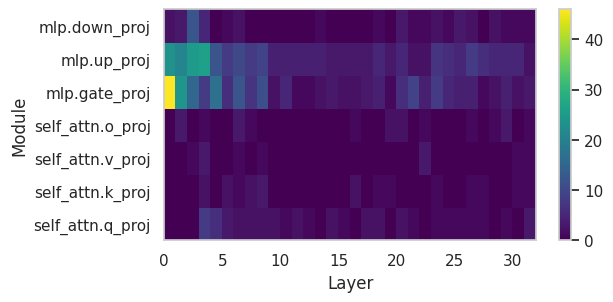}\\
Bold \\
\includegraphics[width=\WidRatio\linewidth]{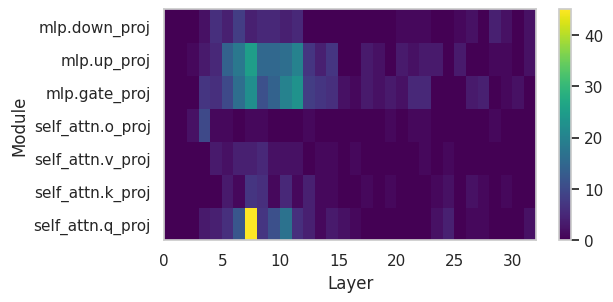}\\
\caption{Distributions of neurons in FsfairX with top and bottom 500 Spearman's $\rho$.}
\label{fig:dist-2d-neuron-fsfairx}
\end{figure}

\clearpage

\begin{table*}[h]
    \tiny
    \centering
    \begin{tabular}{c|c}
    \toprule
    RM & Response \\
    \midrule
    - & \begin{tabular}{p{12cm}}
Larry Page is an American computer scientist and internet entrepreneur. He is best known as the co-founder of Google, along with Sergey Brin. Page is a pioneer in the field of search engines and has played a significant role in shaping the internet as we know it today.\\
\\
Page was born on March 26, 1973, in East Lansing, Michigan. He earned his bachelor's degree in computer engineering from the University of Michigan and later pursued his master's and Ph.D. in computer science from Stanford University, where he met Sergey Brin.\\
\\
In 1998, Page and Brin developed a search engine called Backrub, which later became Google. They refined their algorithm and launched Google in 1998. The company quickly gained popularity due to its relevance-based search results and user-friendly interface.\\
\\
Page served as Google's CEO from 2011 to 2015, during which time he oversaw the company's expansion into new areas such as cloud computing, artificial intelligence, and hardware. In 2015, he stepped down as CEO and became the CEO of Alphabet Inc., the parent company of Google.\\
\\
Page is known for his innovative and visionary approach to technology. He has been involved in various projects, including self-driving cars, high-altitude wind power, and moonshot projects through his venture capital firm, X.\\
\\
In 2019, Page took a temporary leave of absence from Alphabet due to health reasons, and Sundar Pichai took over as CEO of both Google and Alphabet. Page is currently listed as the CEO of Alphabet, but his day-to-day involvement with the company is limited.\\
\\
Throughout his career, Page has received numerous awards and honors, including the Marconi Prize, the National Academy of Engineering's Draper Prize, and the induction into the National Inventors Hall of Fame.
\end{tabular}\\
\midrule
GRM & \begin{tabular}{p{12cm}}
Larry Page is an American computer scientist, internet entrepreneur, and business magnate. He is best known as the co-founder of Google, along with Sergey Brin, and served as the company's CEO from 2011 to 2015.\\
\\
Page was born on March 26, 1973, in East Lansing, Michigan. He earned a Bachelor's degree in computer engineering from the University of Michigan and a Master's degree in computer science from Stanford University, where he met Sergey Brin.\\
\\
In 1998, Page and Brin developed the search engine Google, which quickly gained popularity and became one of the most widely used search engines in the world. In 2001, they incorporated Google and raised initial funding from investors. Page served as Google's CEO from 2011 to 2015, during which time the company expanded its product offerings, made significant investments in artificial intelligence, and acquired several companies.\\
\\
In 2015, Page and Brin restructured Google into a new parent company called Alphabet Inc., with Page becoming CEO of Alphabet and Sundar Pichai taking over as CEO of Google. Page stepped down as CEO of Alphabet in December 2019 and became the company's Chief Architect, focusing on long-term strategic projects.\\
\\
Throughout his career, Page has been recognized for his innovative thinking, entrepreneurial spirit, and philanthropic efforts. He has been awarded several honors, including the Marconi Prize and the IEEE Masaru Ibuka Consumer Electronics Award.\\
\\
Some interesting facts about Larry Page:\\
\\
1. **PageRank**: Page developed the PageRank algorithm, which is the foundation of Google's search engine rankings.\\
2. **Google X**: Page co-founded Google X, a research and development company focused on developing new technologies, including self-driving cars and contact lenses with built-in cameras.\\
3. **Wing**: Page also co-founded Wing, a drone delivery company that aims to revolutionize logistics and transportation.\\
4. **Philanthropy**: Page has donated millions to various causes, including education, healthcare, and environmental conservation.\\
5. **Innovative thinker**: Page is known for his unconventional approach to business and his willingness to take calculated risks.\\
\\
Overall, Larry Page is a visionary entrepreneur, innovator, and business leader who has left a profound impact on the technology industry and the world at large.
\end{tabular}\\
\midrule
GRM+CIRM & \begin{tabular}{p{12cm}}
Larry Page is an American computer scientist, internet entrepreneur, and business magnate. He is best known as the co-founder of Google, along with Sergey Brin, and served as the company's CEO from 2011 to 2015.\\
\\
Page was born on March 26, 1973, in East Lansing, Michigan. He earned a Bachelor's degree in computer engineering from the University of Michigan and a Master's degree in computer science from Stanford University, where he met Sergey Brin.\\
\\
In 1998, Page and Brin developed the search engine Google, which quickly gained popularity and became one of the most widely used search engines in the world. In 2001, they incorporated Google and raised initial funding from investors. Page served as Google's CEO from 2011 to 2015, during which time the company expanded its product offerings, made significant investments in artificial intelligence, and acquired several companies.\\
\\
In 2015, Page and Brin restructured Google into a new parent company called Alphabet Inc., with Page becoming CEO of Alphabet. He held this position until December 2019, when he stepped down and became the CEO of Alphabet's "Other Bets" division, which includes experimental projects such as Waymo (self-driving cars), Verily (life sciences), and X (research and development).\\
\\
Page is known for his innovative thinking, vision for the future of technology, and emphasis on the importance of "moonshot" projects. He has been recognized with numerous awards, including the Marconi Prize and the National Academy of Engineering's Draper Prize.\\
\\
In his personal life, Page is married to Lucinda Southworth, a researcher and sister of actress Elizabeth Southworth. He is a private individual and keeps a relatively low public profile, but his impact on the technology industry and the world at large is undeniable.\\
\\
In 2021, Page stepped down as CEO of Alphabet's "Other Bets" division and was replaced by Jeff Dean, a long-time Google executive. Page continues to serve as a member of Alphabet's board of directors and holds a significant amount of Google's class B shares, giving him significant influence over the company's direction.
\end{tabular} \\
    \bottomrule
    \end{tabular}
    \caption{Examples of responses generated by Llama3 to the prompt ``Who is Larry Page?''.}
    \label{tab:example-gemma2}
\end{table*}

\clearpage

\begin{table*}[h]
    \tiny
    \centering
    \begin{tabular}{c|c}
    \toprule
    RM & Response \\
    \midrule
    - & \begin{tabular}{p{12cm}}
    Larry Page is an American computer scientist and internet entrepreneur. He is best known as the **co-founder of Google** alongside Sergey Brin.\\ 
\\
Here are some key facts about him:\\
\\
* **Born:** March 26, 1973, in East Lansing, Michigan\\
* **Education:** University of Michigan (Bachelor of Science in computer engineering), Stanford University (Master of Science in computer science)\\
* **Key Accomplishments:**\\
~~~~* Co-founded Google in 1998\\
~~~~* Served as Google's CEO from 1998 to 2001 and again from 2011 to 2015\\
~~~~* Led Google's development into a global technology giant\\
~~~~* Founded Alphabet Inc. in 2015, Google's parent company, and served as its CEO\\
~~~~* Known for his innovative ideas and ambitious vision\\
* **Personal Interests:** Aerospace engineering, clean energy, and philanthropy\\
\\
Larry Page played a pivotal role in shaping the internet as we know it today. His creation of Google revolutionized the way people access information and has had a profound impact on countless aspects of our lives.\\
\\
\\
If you'd like to know more about a specific aspect of Larry Page's life or work, feel free to ask!
    \end{tabular}\\
    \midrule
GRM & \begin{tabular}{p{12cm}}
Larry Page is a renowned American computer scientist and internet entrepreneur. Here's a summary of his key achievements and roles:\\
\\
**Co-founder of Google:**\\
\\
* Along with Sergey Brin, Page co-founded Google in 1998.\\
* He developed the PageRank algorithm, a revolutionary system for ranking websites based on their importance and relevance, which became the foundation of Google's search engine.\\
\\
**Leadership Roles:**\\
\\
* **CEO (1998-2001, 2011-2015):**  Page served as Google's CEO twice, leading the company through its explosive growth and expansion into various markets.\\
* **President of Products (2001-2011):**  He focused on product development and innovation during this period.\\
* **Alphabet Inc. CEO (2015-2019):** After restructuring Google into Alphabet Inc., Page became the CEO of the parent company, overseeing a portfolio of diverse businesses including Google, Waymo, Verily, and Calico.\\
\\
**Technological Visionary:**\\
\\
* Page is known for his ambitious vision and relentless pursuit of technological advancements.\\
* He has championed projects like self-driving cars (Waymo), life sciences research (Verily), and artificial intelligence (DeepMind).\\
\\
**Philanthropy:**\\
\\
* Page and Brin have made significant philanthropic contributions through the Google.org foundation and other initiatives.\\
\\
**Current Status:**\\
\\
* While no longer holding an executive position at Alphabet, Page remains involved in the company's strategic direction.\\
\\
Larry Page's contributions have fundamentally shaped the internet landscape and continue to impact our lives in profound ways.
\end{tabular}\\
\midrule
GRM+CIRM & \begin{tabular}{p{12cm}}
**Larry Page** is an American computer scientist and internet entrepreneur. He is best known as the **co-founder of Google** alongside Sergey Brin.\\
\\
Here's a summary of his notable achievements and contributions:\\
\\
* **Google Co-founder:** In 1998, Page and Brin developed the PageRank algorithm, which revolutionized search engine technology. This led to the creation of Google, which quickly became the world's dominant search engine.\\
* **CEO \& President of Google:** Page served as Google's CEO from 1998 to 2001 and again from 2011 to 2015. During his tenure, Google expanded significantly, launching products like Gmail, Android, and Google Maps.\\
* **Alphabet Inc. Founder:** In 2015, Page and Brin restructured Google into Alphabet Inc., a holding company that encompasses Google and other ventures. Page became Alphabet's CEO.\\
* **Innovator \& Visionary:** Page is known for his ambitious vision and drive to push technological boundaries. He has been involved in various projects exploring innovative areas like self-driving cars (Waymo), artificial intelligence (DeepMind), and renewable energy.\\
\\
Larry Page's contributions have had a profound impact on the way we access information, communicate, and interact with the world. He is considered one of the most influential figures in the tech industry.\\
\\
You can find more detailed information about Larry Page on websites like:\\
\\
* **Wikipedia:** https://en.wikipedia.org/wiki/Larry\_Page\\
* **Google:** https://about.google/leadership/larry-page/
\end{tabular}\\
\bottomrule
    \end{tabular}
    \caption{Examples of responses generated by Gemma2 to the prompt ``Who is Larry Page?''.}
    \label{tab:example-llama3}
\end{table*}

\clearpage

\begin{table*}[t]
    \centering
    \begin{tabular}{p{100pt}|p{300pt}}
    \multicolumn{2}{c}{\textbf{Reward Model}}\\
    \toprule
    \textbf{Model} & \textbf{HuggingFace Link} \\
    \midrule
    FsfairX & \href{https://huggingface.co/sfairXC/FsfairX-LLaMA3-RM-v0.1}{\texttt{sfairXC/FsfairX-LLaMA3-RM-v0.1}} \\
    \rowcolor[gray]{\RowColor}
    GRM & \href{https://huggingface.co/Ray2333/GRM-gemma2-2B-rewardmodel-ft}{\texttt{Ray2333/GRM-gemma2-2B-rewardmodel-ft}} \\
    INF & \href{https://huggingface.co/infly/INF-ORM-Llama3.1-70B}{\texttt{infly/INF-ORM-Llama3.1-70B}}\\
    \bottomrule
    \end{tabular}
    \vspace{7pt}

    \begin{tabular}{p{100pt}|p{300pt}}
    \multicolumn{2}{c}{\textbf{Benchmark for Reward Model}}\\
    \toprule
    \textbf{Data} & \textbf{HuggingFace Link} \\
    \midrule
    RewardBench & \href{https://huggingface.co/datasets/allenai/reward-bench}{\texttt{allenai/reward-bench}} \\
    \rowcolor[gray]{\RowColor}
    RM-Bench & \href{https://huggingface.co/datasets/THU-KEG/RM-Bench}{\texttt{THU-KEG/RM-Bench}} \\
    \bottomrule
    \end{tabular}
    \vspace{7pt}

    \begin{tabular}{p{100pt}|p{300pt}}
    \multicolumn{2}{c}{\textbf{Language Model}}\\
    \toprule
    \textbf{Model} & \textbf{HuggingFace Link} \\
    \midrule
    Llama-3-8B-Instruct & \href{https://huggingface.co/meta-llama/Meta-Llama-3-8B-Instruct}{\texttt{meta-llama/Meta-Llama-3-8B-Instruct}} \\
    \rowcolor[gray]{\RowColor}
    Gemma-2-9B-it & \href{https://huggingface.co/google/gemma-2-9b-it}{\texttt{google/gemma-2-9b-it}}\\
    \bottomrule
    \end{tabular}
    \vspace{7pt}

    \begin{tabular}{p{100pt}|p{300pt}}
    \multicolumn{2}{c}{\textbf{On-policy Responses for Preference Annotation}}\\
    \toprule
    \textbf{Data} & \textbf{HuggingFace Link} \\
    \midrule
    Llama-3-8B-Instruct & \href{https://huggingface.co/datasets/princeton-nlp/llama3-ultrafeedback-armorm}{\texttt{princeton-nlp/llama3-ultrafeedback-armorm}} \\
    \rowcolor[gray]{\RowColor}
    Gemma-2-9B-it & \href{https://huggingface.co/datasets/princeton-nlp/gemma2-ultrafeedback-armorm}{\texttt{princeton-nlp/gemma2-ultrafeedback-armorm}} \\
    \bottomrule
    \end{tabular}
    \vspace{7pt}

    \begin{tabular}{p{100pt}|p{300pt}}
    \multicolumn{2}{c}{\textbf{Benchmark for Language Model}}\\
    \toprule
    \textbf{Benchmark} & \textbf{GitHub Link}\\
    \midrule
    AlpacaEval 2.0 & \href{https://github.com/tatsu-lab/alpaca_eval}{\texttt{github.com/tatsu-lab/alpaca\_eval}} \\
    \rowcolor[gray]{\RowColor}
    MT-Bench & \href{https://github.com/lm-sys/FastChat/tree/main/fastchat/llm_judge}{\texttt{github.com/lm-sys/FastChat/tree/main/fastchat/llm\_judge}} \\
    TruthfulQA & \href{https://github.com/EleutherAI/lm-evaluation-harness}{\texttt{github.com/EleutherAI/lm-evaluation-harness}} \\
    \bottomrule
    \end{tabular}
    \caption{Models and datasets we used in our experiments.}
    \label{tab:resources}
\end{table*}

\end{document}